\documentclass[runningheads]{llncs}
\usepackage{graphicx}

\usepackage{longtable}
\usepackage{multirow}
\usepackage{nameref}
\usepackage{booktabs}
\usepackage{pifont}
\usepackage{changepage}
\usepackage{listings}
\usepackage{url}
\usepackage{xspace}
\usepackage{xtab}
\usepackage{float}
\usepackage{cite}
\usepackage{epsfig}           
\usepackage{subfigure}
\usepackage[dvipsnames]{xcolor}
\begin{document}

\title{End-to-End Approach for Recognition of Historical Digit Strings}
%
%
\author{Mengqiao Zhao\inst{1}, Andre G. Hochuli\inst{2}, Abbas Cheddad\inst{1}}
%
%
\institute{
Faculty of Computing, Blekinge Institute of Technology, 371 79 Karlskrona, Sweden\\
\email{\{mengqiao.zhao, abbas.cheddad\}@bth.se}
\vspace{3mm}	
\and
Pontifical Catholic University of Parana (PPGIa/PUCPR), Curitiba, Brazil \\
R. Imaculada Concei\c c\~ao, 1155, Curitiba, PR, Brazil - 80215-901 \\
\email{aghochuli@ppgia.pucpr.pr}
}
\maketitle              
\begin{abstract}
		The plethora of digitalised historical document datasets released in recent years has rekindled interest in advancing the field of handwriting pattern recognition. In the same vein, a recently published data set, known as ARDIS, presents handwritten digits manually cropped from 15.000 scanned documents of Swedish church books and exhibiting various handwriting styles. To this end, we propose an end-to-end segmentation- free deep learning approach to handle this challenging ancient handwriting style of dates present in the ARDIS dataset (4-digits long strings). We show that with slight modifications in the VGG-16 deep model, the framework can achieve a recognition rate of 93.2\%, resulting in a feasible solution free of heuristic methods, segmentation, and fusion methods. Moreover, the proposed approach outperforms the well-known CRNN method (a model widely applied in handwriting recognition tasks).\textcolor{blue}{[Cite as: Mengqiao Zhao, Andre G. Hochuli and Abbas Cheddad, End-to-End  Approach for Recognition of Historical  Digit  Strings,  to appear  in  the  16th  International  Conference  on  Document Analysis  and  Recognition (ICDAR 2021), LNCS, Springer, Lausanne, Switzerland.]}

\end{abstract}
\section{Introduction}
Due to the rapid growth of document storage in modern society, handwriting recognition has become an important research branch in the field of machine learning and pattern recognition. In the context of handwriting digit strings recognition (HDSR), there are several application scenarios such as postcode recognition, bank checks, document indexing\cite{cecotti2016active} and word spotting \cite{Almazan2014,Cheddad16,Gao2020}. 

Most state-of-the-art approaches focus on the segmentation of connected components, followed by training a model to classify each component. However, the performance collapses when two or more digits are touching. Besides, variant approaches based on multiple segmentation algorithms have been proposed. For such algorithms, to generate a potential segmentation cut, a heuristic analysis based on background and foreground information, contour, shape, or a combination of these is required to be implemented\cite{ribas2013handwritten}. The over-segmentation strategy is frequently used to optimize the segmentation of touching components by over segmenting the string. Although over-segmentation increases the probability of obtaining a plausible segmentation cut, it also increases the computational cost since the hypothesis space expands exponentially with the increase of the segmentation cuts. However, the use of heuristic to deal with touching digits has shown that performance degrades by the presence of noise, fragments, and the lack of context in digit strings, such as a lexicon and the string length. 

To better address these issues and to take advantage of deep learning models, segmentation-free methods came to the surface, offering unique capabilities to the research community in the domain \cite{hochuli2018handwritten,Aly2019,Hochuli2020,Flor2020}. Along this line, the proposed approaches rely on implicit segmentation through a deep model or a set of them. Recently, Hochuli et al.\cite{Hochuli2020b} evaluated object recognition models into the HSDR context. For that, a digit string is considered a sequence of objects. The advantage is that these models can encode the background, the shape, and the neighbourhood of digits efficiently. Nonetheless, the annotation of digit bounding boxes (ground-truth) is a drawback when synthetic data are not available. Finally, sequence-to-sequence approaches were proposed resulting in feasible end-to-end models \cite{shi2016end} for word strings’ recognition. The primary strategy is to split the input string into fragments to feed a recurrent model (RNN), and then, a transcription layer determines the resulting string. As stated in \cite{Flor2020,Hochuli2020b}, due to the lack of context, this approach did not achieve outstanding results in HDSR; however, it produces a good trade-off between data annotation, training complexity and accuracy.

An essential aspect of the strategies mentioned above is that most of them have been proposed for modern handwritten digit strings recognition. There is still a lack of approaches in the context of historical (ancient) document recognition.The challenges are different from modern ones, such as paper texture deterioration, noise, ancient handwriting style, ink failure, bleed-through, and the lack of data \cite{kusetogullari2019ardis}. Remarkably, the performance of the modern approaches applied to historical document context is a matter of discussion. 

Recently, Kusetogullari et al.\cite{kusetogullari2019ardis} released to the public the ARDIS dataset, composed of historical handwriting digit strings extracted from the Swedish church record. Their comprehensive analysis reveals that a model trained with modern isolated digits (MNIST, USPS, etc.) fails by a fair margin to correctly encode the isolated digit from ARDIS due to its unique characteristics. However, a comprehensive analysis using the historical digit strings of ARDIS dataset is missing.
In light of this, we propose to assess two state-of-the-art approaches by adapting slight modifications into their architecture to better fit the ARDIS digit strings characteristics. Moreover, we introduce data augmentation techniques to represent the classes more efficiently. This work’s resulting analysis eventually proposes a baseline for the ARDIS strings dataset and pin-points, which efforts are needed to implement feasible solutions for historical digit strings recognition. Moreover, it highlights the research gaps for further investigations.

The rest of this document is organized as follows: The related work is presented in Section \ref{sec:Related Work}. The assessed approaches are described in Section \ref{sec:approach}, and then Section \ref{sec:Experiments} provides a discussion around the experiments. Section \ref{sec:conclusion} presents the conclusion and future directions.

\section{Related Work}
\label{sec:Related Work}

We surveyed the state-of-art and divided the related work into two main categories: (a) Segmentation-based and (b) Segmentation-free approaches. The related work is discussed in the following paragraphs.

\noindent \textbf{Segmentation-based approaches}: By segmenting the connected components as much as possible, we attain the concept of over-segmentation, which is the most commonly used strategy. It maximizes the probability of generating the optimal segmentation point. However, as mentioned earlier, it also increases the hypothesis space, resulting in a higher computational cost to classify all the candidates compared with a strategy based on single segmentation.
	
An implicit filter is proposed by Vellasques et al. \cite{vellasques2008filtering} to reduce the computational cost of over-segmentation,in which a Support Vector Machine (SVM) classifier is used to determine whether the cut produces reliable candidates. The proposed filter succeeded in eliminating up to 83\% of unnecessary segmentation cuts in their experimental results.

In Roy et al. ~\cite{RoyVajda2005}, a segmentation-based approach is devised to segment out destination address block for postal applications; a review of postal and check processing applications is warranted in ~\cite{gilloux2014}.

Besides the strategies mentioned above, several segmentation algorithms were proposed in the last decade. Ribas et al. \cite{ribas2013handwritten} assessed most of them considering their performance, computational cost, touching types, and complexity. This characterization aimed to identify the limitations of the algorithms based on a given pair of touching digits. Moreover, the work reveals that most of the heuristic segmentation strategies are biased towards the characteristic of the dataset’s characteristic under scrutiny; thus, a suitable method that works for all touching types is impractical.
	
\noindent\textbf{Segmentation-free approaches}: To the best of our knowledge, the first attempt along this line was proposed by Matan et al. \cite{matan1992multi}. A convolutional neural network (CNN) based model is displaced from left to right over the input. The proposed approach is termed SDNN (Space Displacement Neural Network), which reported 66\% of correct classification on 3000 images of ZIP Codes. LeCun et al. \cite{lecun1998gradient}, stated that SDNN is an attractive technique but has not yielded better results than heuristic segmentation methods.

Years later, Choi and Oh \cite{choi2001segmentation} presented a modular neural network composed of 100 sub-networks trained to recognize 100 classes of touching digits (00..99). The recognition rate of 1374 pairs of digits extracted from the NIST database reaches 95.3\%. A similar concept was presented by Ciresan \cite{ciresan2008avoiding}, in which 100-class CNN was trained with 200,000 images reporting a recognition rate of 94.65\%. 
	
An image-based sequence recognition was proposed by Shi et al.\cite{shi2017}. The end-to-end framework, named Convolutional Recurrent Neural Network (CRNN), naturally handles sequences in arbitrary lengths without character segmentation or horizontal scale normalization. The approach achieved outstanding performance on recognising the scene text (text in the wild) and music scores.	
	
To make handwriting digit recognition less dependent on segmentation algorithms, Hochuli et  al.\cite{hochuli2018handwritten} proposed a segmentation-free framework based on a dynamic selection of classifiers. The authors postulate that a set of convolutional neural networks trained to (a) predict the size of touching components and (b) specific-task models to recognize up to three touching digits performs better than if the digits were segmented. However, this algorithm's generalisation to other datasets needs further verification since there is a lack of diversity of the used datasets in the experimental protocol.
	
Cheng et al. \cite{cheng2019handwritten} proposed a strategy based on the improved VGG-16 model to overcome the lack of texture features in handwriting digit recognition. The model was examined on the extended MNIST dataset, eventually achieved a high accuracy of 99.97\%, which indicates that this VGG-based model has a robust feature extraction ability than traditional classifiers and can meet the requirements of handwriting digits classification and recognition. Besides, a VGG-like model with multiple sub classifiers was built to recognize CAPTCHAs. Although the CAPTCHA images for the test are featured by a lot of noise and touching digits, the model accuracy reached 98.26\% without any pre-segmentation.

End-to-end approaches are frequently proposed in recent years, including those tackling writer identification ~\cite{CILIA2020137} and document analysis and recognition ~\cite{Neudecker2019, Palm2019}.
Recently, approaches based on object recognition models have been exploited with the HDSR task \cite{Hochuli2020,Hochuli2020b}. Considering that a string is a sequence of objects, these models can efficiently encode the background, shape, and neighbourhood of digits, providing an end-to-end solution for the problem. Additionally, they reduce the restrictions imposed on the number of touching components or string length. However, the annotation of each digit bounding box (ground-truth) is a bottleneck when synthetic data are not applicable.

\section{ARDIS dataset}\label{sec:ardis}

\begin{figure}[!ht] 
	\centering 
	\includegraphics[width=.65\textwidth]{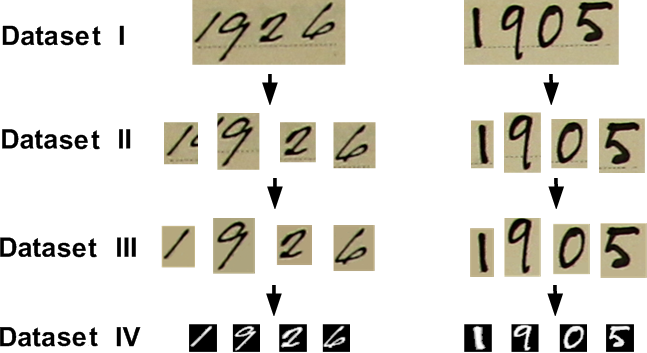} 
	\caption{Examples from the different representations of ARDIS subsets \cite{kusetogullari2019ardis}.}
	\label{fig:ardis_dataset} 
\end{figure}

The Arkiv Digital Sweden (ARDIS) Dataset comprises historical handwriting digit strings extracted from Swedish church document images written by different priests from 1895 to 1970. The dataset is fully annotated \cite{kusetogullari2019ardis}, including the digits bounding-boxes \cite{Hochuli2020}. The sub-datasets of ARDIS are exemplified in Figure \ref{fig:ardis_dataset}. 

\begin{figure}[!ht] 
	\centering 
	\includegraphics[width=.6\textwidth]{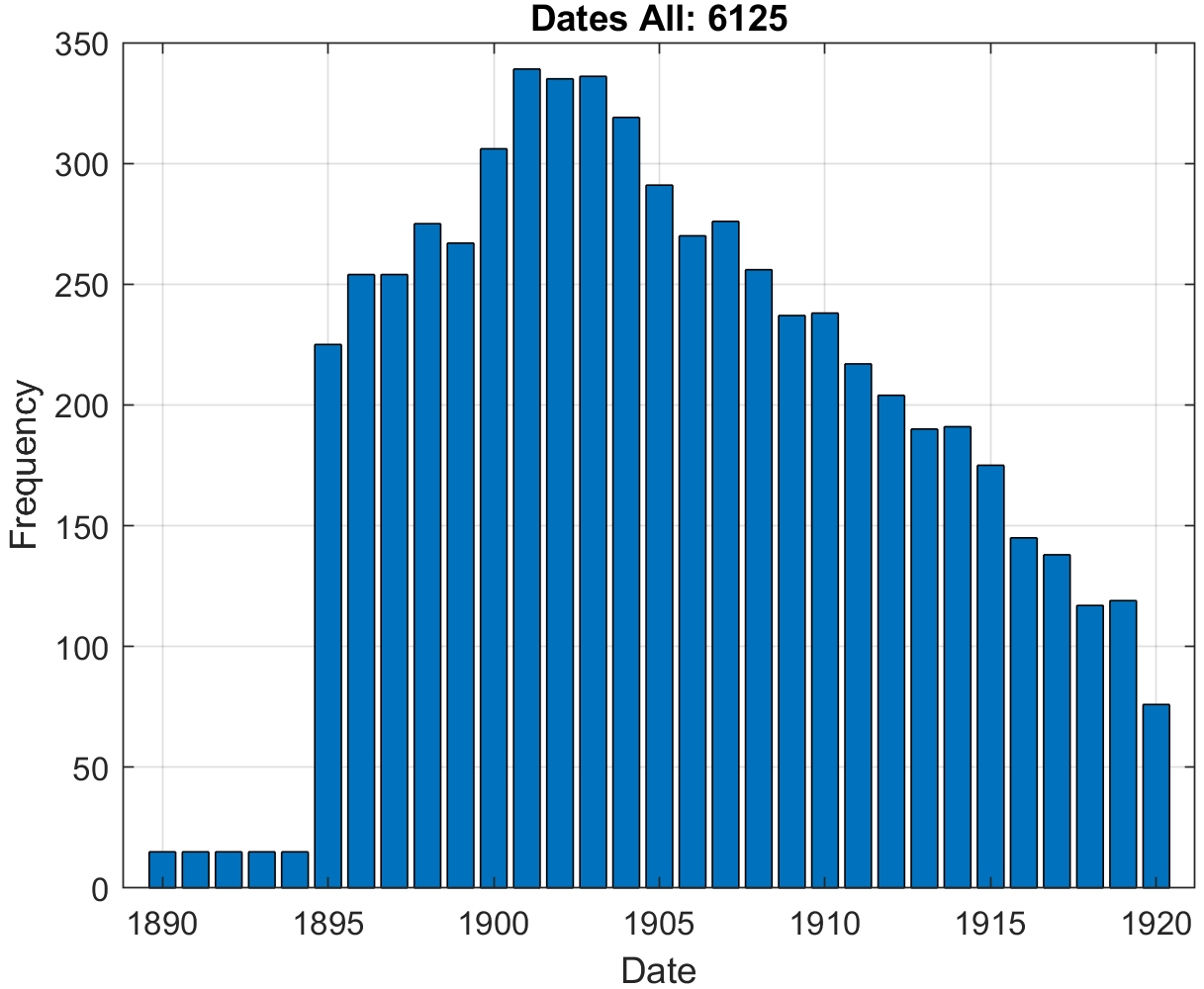} 
	\caption{Class distribution of ARDIS Strings\cite{kusetogullari2019ardis}. Classes later than 1920 were not considered in this work due to an insufficient number of samples.}
	\label{fig:ardis_distribution} 
\end{figure}

The dataset (I) is composed of 4-digit strings that represent the year of a record. Most of the samples were cropped with the size of 175 x 95 pixels from the document image and stored in its pristine RGB colour space. The dataset (I) contains 75 classes mapping to different years. However, due to insufficient samples, classes later than 1920 were not considered in this work. The class distribution used in this study is shown in Figure \ref{fig:ardis_distribution}.

\begin{figure}[!ht] 
	\centering 
	\includegraphics[width=0.75\textwidth]{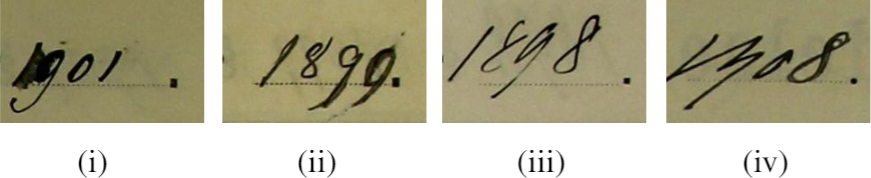} 
	\caption{Challenging samples in ARDIS Dataset (I) representing noisy handwritten digits (i)(ii), ink failure (iii) and touching digits (iv). 
	}
	\label{fig:ardis_challenges} 
\end{figure}

The historical digit strings pose several challenges to classification, including  variations in terms of variability of handwriting styles, touching digits, ink failure, and noisy handwriting. Figure \ref{fig:ardis_challenges} demonstrates some of the aforementioned challenges. Moreover, as observed in Figure \ref{fig:ardis_distribution}, the classes are not equally distributed. 

The dataset (II) comprises cropped digits from the original digit strings, containing artefacts and fragments of the neighbour digits. In dataset (III), the isolated digits were manually cleaned. For completeness, a uniform distribution of each digit's occurrences was ensured in the dataset (III), resulting in 7600 de-noised digit images in RGB colour space. For dataset (IV), the isolated digits are normalised and binarised.

\subsection{Synthetic Data}\label{sec:synthetic_data}

\begin{figure}[!ht] 
	\centering 
	\includegraphics[width=0.6\textwidth]{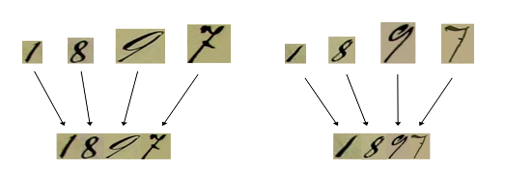} 
	\caption{Synthetic strings creation from isolated digits of dataset III.}
	\label{fig:sythetic_data} 
\end{figure}

Recently, data augmentation techniques in handwriting digit were proposed to generate a synthetic training set \cite{shi2016end,Aly2019, hochuli2018handwritten}. In order to improve the data representation of ARDIS strings (Dataset I), we propose the creation of synthetic data by permuting and concatenating several single digits from dataset III. Figure \ref{fig:sythetic_data} depicts two synthetic samples. Although both digit strings belong to the label ``1987'', the representation (e.g., style) is remarkably different. 

We combine synthetic data and real data from up to 1000 samples for each class during the training phase to balance the distribution of classes. The number of real and synthetic data are summarized in Table \ref{tab:synthetic_data}. Considering the limited amount of data in the ARDIS data set, it is relatively reasonable (closer to standard practice) to retain about 43\% of the real data for testing (2651/(3474+2651)) and use the remaining 57\% of the real data set for training. It is worth mentioning that the synthetic data was used only for training, which represents 88.8\% of the training samples. In total, 31000 and 2651 images are used for training and testing, respectively. All the data used in this work are already publicly available through the ARDIS Website\footnote{https://ardisdataset.github.io/ARDIS/} by the authors of \cite{kusetogullari2019ardis}.

\begin{table}[!ht]
	\centering
	\caption{Training protocol using synthetic and real data}
	\resizebox{0.5\textwidth}{!}{%
	\begin{tabular}{cccc}
		\hline
		\multirow{2}{*}{\textbf{Label  }} & \multicolumn{2}{c|}{ \textbf{   Training Samples   } } & \multirow{2}{*}{\textbf{   Test Samples   }}\cr\cline{2-3} 
		& \textbf{   Synthetic  } & \textbf{  Real  } & \cr 
		\hline
		1890 & 1000 & 0 & 15 \\ 
		1891 & 1000 & 0 & 15 \\ 
		1892 & 1000 & 0 & 15 \\ 
		1893 & 1000 & 0 & 15 \\ 
		1894 & 1000 & 0 & 15\\ 
		1895 & 875 & 125 & 100 \\ 
		1896 & 846 & 154 & 100 \\ 
		1897 & 846 & 154 & 100 \\ 
		1898 & 825 & 175 & 100 \\ 
		1899 & 833 & 167 & 100 \\ 
		1900 & 794 & 206 & 100 \\ 
		1901 & 761 & 239 & 100  \\ 
		1902 & 765 & 235 & 100 \\ 
		1903 & 764 & 236 & 100 \\ 
		1904 & 781 & 219 & 100 \\ 
		1905 & 809 & 191 & 100 \\ 
		1906 & 830 & 170 & 100 \\ 
		1907 & 824 & 176 & 100 \\ 
		1908 & 844 & 156 & 100 \\ 
		1909 & 863 & 137 & 100 \\ 
		1910 & 862 & 138 & 100 \\ 
		1911 & 883 & 117 & 100 \\ 
		1912 & 896 & 104 & 100 \\ 
		1913 & 910 & 90 & 100 \\ 
		1914 & 909 & 91 & 100 \\ 
		1915 & 925 & 75 & 100 \\ 
		1916 & 955 & 45 & 100 \\ 
		1917 & 962 & 38 & 100 \\ 
		1918 & 983 & 17 & 100 \\ 
		1919 & 981 & 19 & 100 \\ 
		1920 & 1000 & 0  & 76 \\ \hline
		\textbf{Total} & 27526 & 3474 & 2651 \\ \hline
	\end{tabular}%
	}
	\label{tab:synthetic_data}
\end{table}

\section{Approaches for Historical Handwriting Digit String Recognition}\label{sec:approach}

As stated by \cite{ribas2013handwritten} and \cite{hochuli2018handwritten}, the segmentation problem has been overcome by segmentation-free approaches in the recent advances in deep learning models. In light of this, we propose to evaluate two segmentation-free approaches on the ARDIS dataset to tackle the task of historical handwriting digit string recognition. The first approach (Section \ref{sec:specific-task}) is based on the well-known VGG-16 model. The second one (Section \ref{sec:crnn}) is based on a sequence-to-sequence model. 

\textit{Motivations on the choice of models:} Given the VGG-16 model's decent performance in other character recognition tasks, we consider it the baseline of this experiment to evaluate against other alternative models. The VGG network model \cite{simonyan2015very} was proposed by the Visual Geometry Group (VGG) at Oxford University. When first created, the focus of this network was to classify materials by their textural appearance and not by their colour. Due to the excellent generalisation performance of VGG-Net, its pre-trained model on the ImageNet dataset is widely used for feature extraction problems \cite{cimpoi2015deep,gao2015deep} such as: object candidate frame (object proposal) generation \cite{ghodrati2017deepproposal}, fine-grained object localization, image retrieval \cite{wei2017selective}, image co-localization \cite{wen2016learning}, etc. On the other hand, our new approach is based on modifying the concept of CRNN \cite{shi2017}. The CRNN is mainly used for end-to-end recognition of indefinite length text sequence. It does not require pre-segmentation on long continuous text.

\subsection{Specific-Task Classifiers}\label{sec:specific-task}

\begin{figure}[!ht] 
 	\centering 
 	\includegraphics[width=0.25\textwidth]{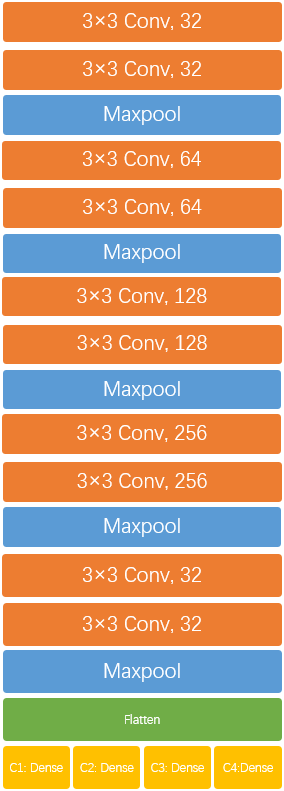} 
 	\caption{Structure of the 4 Classifiers CNN framework}
 	\label{fig:4C} 
 \end{figure}

 In this new approach, we adapted the well-known VGG-16 model to the context of the ARDIS dataset, which is composed of 4-digit strings. Instead of using a dynamic selection of classifiers \cite{hochuli2018handwritten}, we proposed to parallelize the classification task by adding four dense layers (classifiers) on the bottom of the architecture. The final architecture is depicted in Figure \ref{fig:4C}. With this simple modification, we produce an end-to-end pipeline avoiding both the heuristic segmentation and fusion methods. 
 
 The rationale here is that each specific-task classifier ($C_{1}$, $C_{2}$, $C_{3}$, and $C_{4}$) should determine the ten classes (0..9) for each digit of the 4-digit string. The prediction of input digit string is defined as follows:

  Let $\mathcal{C}(x) = \max\limits_{0 \leq i \leq 9} p^i(x)$ be the probability produced by the digit classifier (10-classes).  Then, an input digit $x$ is assigned to the class $\omega_j$ (j={0...9}) according to Equation \ref{eq:prob1}.
 
  \begin{equation}
   \label{eq:prob1}
	P(\omega_j|x) = \max(\mathcal{C})
   \end{equation}
	
  Considering that the input image $I$ contains $n = 4$ digits, the most probable interpretation of the written amount $M$ is given by Equation \ref{eq:prob2}.  
 \begin{equation}
 \label{eq:prob2}
	 P(M|I) = \prod_{i=1}^n P_{i}(\omega_j|x_i)
 \end{equation}

\subsection{CRNN}\label{sec:crnn}

\begin{figure}[h!]
	\centering
	\subfigure[] {\epsfig {file=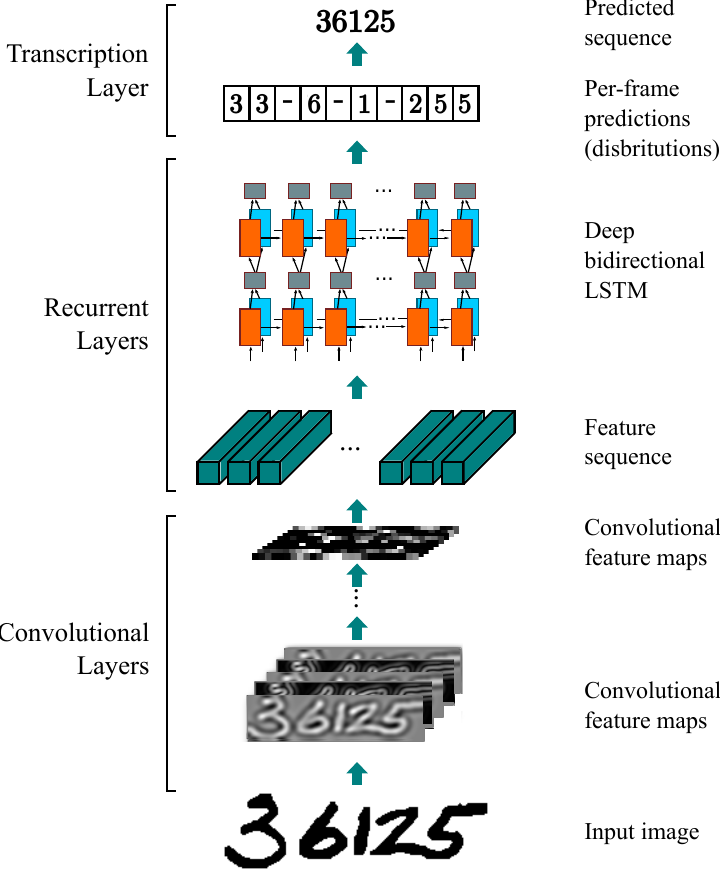, width=.55\textwidth}}
	\hspace{1cm}
	\subfigure[] {\epsfig {file=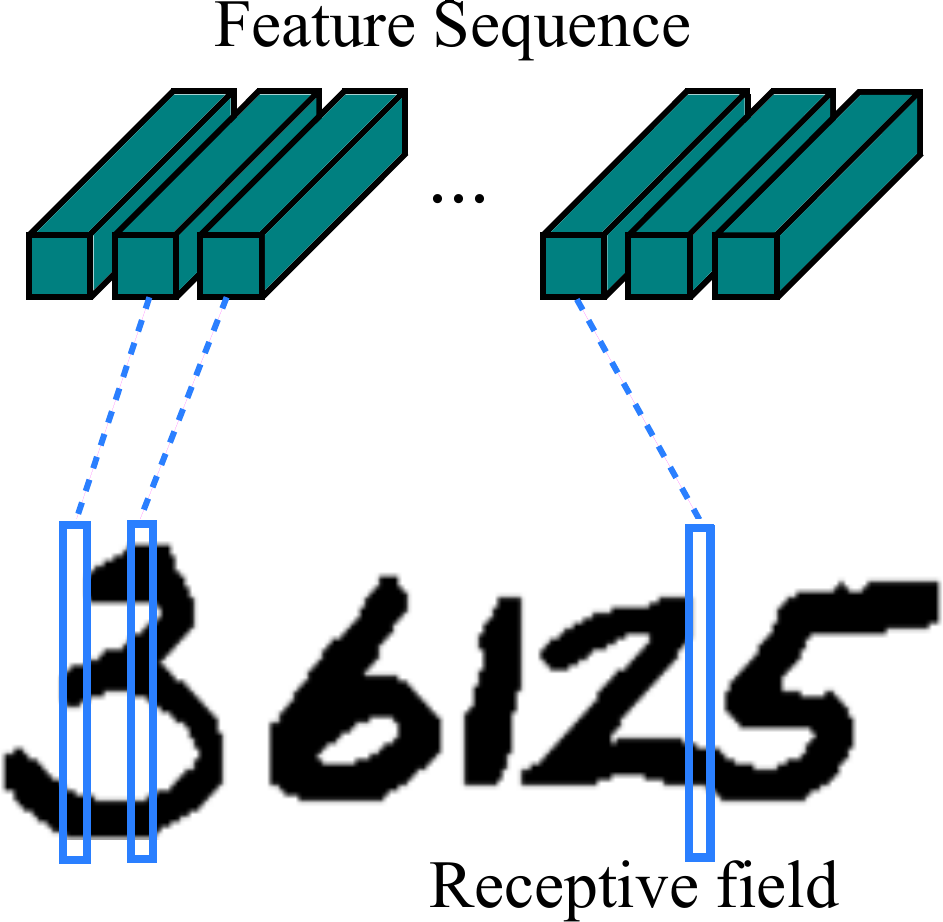, width=.3\textwidth}}
	\caption{CRNN architecture proposed by \cite{shi2016end}: (a) the pipeline from convolutional layers to transcription layer and (b) the receptive field for each feature vector.}
	\label{crnn_pipe:fig}
\end{figure}

A Convolutional Recurrent Neural Networks (CRNN) \cite{Voigtlaender2016,shi2016end,Dutta2018} is a sequence-to-sequence model that can be trained from end to end. The pipeline for such a network is depicted in Figure \ref{crnn_pipe:fig}a. First, convolutional layers extract features from an input image, and then a sequence of feature vectors is extracted from feature maps.

Since each region of the feature map is associated with a receptive field in the input image, each vector in the sequence is a descriptor of this image field, as illustrated in Figure \ref{crnn_pipe:fig}b. Next, this sequence is fed to the recurrent layers, which are composed of a bidirectional Long-Short Term Memory (LSTM) \cite{Schuster1997} network, producing a per-frame prediction from left to right of the image. Finally, the transcription layer determines the correct sequence of classes to the input image by removing the repeated adjacent labels and the blanks, represented by the character `-'. This solution is well suited when the past and future context of a sequence contributes to recognising the whole input. With the aid of contextual information, such as a lexicon, this approach achieves high text recognition performance. The application of this solution to handwriting digits is a matter of discussion since we have fewer classes (0..9) as compared to words, but there is no lexicon to mitigate possible confusion.

\begin{figure}[!ht] 
	\centering 
	\includegraphics[width=0.6\textwidth]{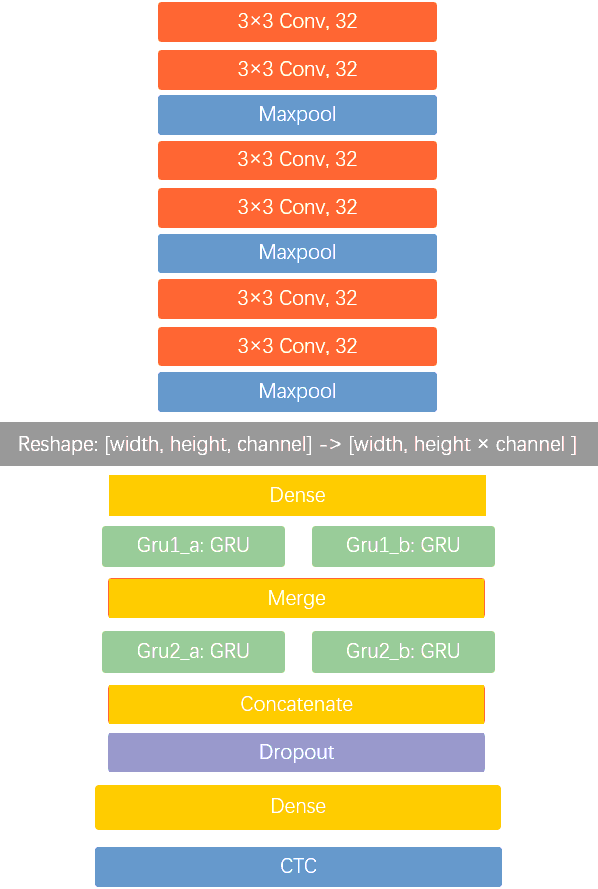} 
	\caption{CRNN architecture for HDSR.}
	\label{fig:crnn_hhdsr} 
\end{figure}

To address the context of historical digit string recognition, we propose a modification of the Recurrent Layers Architecture. Due to the lack of data, we replaced the LSTM with a Gated Recurrent Unit (GRU) \cite{cho2014learning}. Since the latter has fewer parameters, besides reducing training time, the vanishing gradient's impact is minimised. Moreover, we combined two identical GRU Layers to process the feature maps from left to right and vice-versa, and then, the output of both GRUs are merged. It is worth mentioning that the feature maps fed to the GRU are reshaped to vectors to provide a sequence of information. Further, another two identical GRUs repeat the process; however, their outputs are concatenated. Finally, a fully connected layer determines the class probabilities, and the connectionist temporal classification (CTC) layer determines the final prediction. The architecture and characterisation of our modified CRNN approach are depicted in Figure  \ref{fig:crnn_hhdsr}.

\section{Experiments}
\label{sec:Experiments}

	In this section, we assess all reported models in the context of HDSR. We also added to the comparison the native VGG16 model (i.e., without our modification). All metrics used to measure the performance are described in Section \ref{sec:metrics}. The training protocol is presented in Section \ref{sec:training}. Finally, the results are discussed in Section \ref{sec:discussion}.
	
\subsection{Evaluation Metrics}\label{sec:metrics}

To better assess the performance of the proposed approaches, besides the well-known accuracy and F1-score, we propose using the Normalized Levenshtein Distance (NLD) and the Average Normalized Levenshtein Distance (ANLD). 

The Normalized Levenshtein Distance (NLD) \cite{diem2014icfhr} describes how close a predicted string is from the ground truth by eliminating the influence of string length on performance measurement. The NLD can be defined as follow:

\begin{equation}
NLD(a_{T},a_{R})= \frac{LD(a_{T},a_{R})}{\left | a_{T} \right |}
\end{equation}

\noindent where \(\left | a_{T} \right |\) presents the length of the string \(a_{T}\) and LD represents the Levenshtein distance, which refers to the minimum number of editing operations of each character (insert, delete, and substitute) to convert from one string to another. A zero value indicates a correct prediction.

Complementary to this, the Average Normalized Levenshtein Distance (ANLD) is a soft metric to evaluate the model performance:
\begin{equation}
ANLD = \frac{\sum_{i=1}^{T}NLD\left ( a_{T}^{i}, a_{R}^{i} \right )}{T}
\end{equation}

\noindent where T indicates the number of evaluated strings. Lower values represent a good performance. 

\subsection{Training}\label{sec:training}

The models were fine-tuned using the data described in Table \ref{tab:synthetic_data}, which comprises real and synthetic data that sums up to 1000 training samples for each one of 31 classes (number of years in the selected period). All the classifiers were trained up to 100 epochs, and the over-fitting was prevented through early-stopping when no convergence occurs after ten epochs. The training parameters are summarized in  Table \ref{tab:hyperparameters}.

\begin{table}[!ht]
	\caption{Training Parameters}
	\label{tab:hyperparameters}
	\begin{tabular}{ccccccc}
		\hline 
		\multirow{2}{*}{\textbf{Frameworks}} & \textbf{Batch} & \textbf{Loss} & \multirow{2}{*}{\textbf{Reg.}} & \multirow{2}{*}{\textbf{Opt.}} & \textbf{Learning} & \textbf{Batch} \\
		& \textbf{Size} & \textbf{Function} &  &  & \textbf{Rate} & \textbf{Norm} \\ \hline
		\textbf{Specific-Task} & 32 & Cat. Crossentropy & Dropout (0.25) & Adam & $10^{-3}$ & TRUE \\
		\textbf{CRNN} & 128 & CTC & Dropout (0.25) & Adadelta & $10^{-3}$ & FALSE \\
		\textbf{VGG-16} & 32 & Cat. Crossentropy & L2 ($5*10^{-4}$) & Adam & $10^{-5}$ & FALSE \\
		\hline
	\end{tabular}
\end{table}

\subsection{Results and Discussion}\label{sec:discussion}

 The performance of the evaluated models in this work is reported in Table \ref{tab:comparison}. As described in Section \ref{sec:specific-task}, we modified the last layer of VGG-16 architecture by adding four classifiers instead of one. As stated in Table \ref{tab:comparison}, this modification achieved the best performance since it can encode the string by implicit segmenting the digits with the specific-task classifiers. Comprehensive analysis reveals that the whole image's information is difficult to encode in several cases by only one classifier (VGG-16) due to the handwriting variability. For example, let us assume that we have the following strings “1890” and “1819”. From a computational perspective, the global representation poses challenges in discriminating the digit ‘0’ and ‘9’ that mislead the classifier. However, for the specific-task approach, the information can be implicitly segmented according to each classifier's domain space. Regarding this, the models  $C_{1}$ and $C_{2}$ exhibit a reduced complexity when compared to the models $C_{3}$ and $C_{4}$ since the former two classifiers need to discriminate fewer classes ([1],[8,9]). 
 
  \begin{table}[h!]
 	\centering
 	\caption{Comparisons among the three frameworks in terms of different metrics}
 	\label{tab:comparison}
 	
 	\begin{tabular}{cccc}
 		\hline
 		\textbf{Frameworks} & \textbf{Accuracy} & \textbf{F1-score}& \textbf{ANLD}\\ \hline
 	VGG-16 & 36.2 & 38.5 & 28.7 \\ 
 	CRNN & 85.0 & 83.5 & 5.8 \\ 
 	\textbf{Specific-Task}  & 93.2 & 90.8 & 2.3 \\ \hline
 \end{tabular}
\end{table}
 
 Regarding the CRNN, we believe that the model suffers due to a lack of context. Contrary to the word recognition, the CTC layer missed the prediction since there is no lexicon to mitigate some confusions, such as fragment recognition and repeated labels. The issue is quite similar to the over-segmentation.

\section{Conclusion and Future Work}\label{sec:conclusion}
In this work, we explored the recognition of historical digit strings. Based on this context, an image-based dataset containing 31 classes representing handwriting years ranging from 1890 to 1920 is utilised.

To this end, we proposed to evaluate three models implementing end-to-end solutions. The use of synthetic data was employed to overcome the lack of data.

The proposed approach that combines four specific-task classifiers achieved outstanding results. This promising performance can be explained by the implicit segmentation of the input string made by the domain space of each specific-task classifier. On the other hand, the approach based on a single classifier suffers due to the handwriting variability represented in a global perspective. Regarding the CRNN approach, it suffers from the lack of lexicon, as also stated in \cite{Hochuli2020b}.

For future endeavours, once we have context information about the first and second digits, a reduced number of classifiers for the specific-task approach could be examined. Also, we will investigate a dynamic approach on VGG-16 that can implicitly handle different length of strings.

\paragraph{ \textbf{Acknowledgments.} \normalfont{This project is supported by the research project “DocPRESERV: Preserving and Processing Historical Document Images with Artificial Intelligence”, STINT, the Swedish Foundation for International Cooperation in Research and Higher Education (Grant: AF2020-8892).}}

\bibliography{ICDAR21Cheddad}

\begin{thebibliography}{10}
\providecommand{\url}[1]{\texttt{#1}}
\providecommand{\urlprefix}{URL }
\providecommand{\doi}[1]{https://doi.org/#1}

\bibitem{Almazan2014}
{Almazan}, J., {Gordo}, A., {Fornés}, A., {Valveny}, E.: Word spotting and
  recognition with embedded attributes. IEEE Transactions on Pattern Analysis
  and Machine Intelligence  \textbf{36}(12),  2552--2566 (2014).
  \doi{10.1109/TPAMI.2014.2339814}

\bibitem{Aly2019}
{Aly}, S., {Mohamed}, A.: Unknown-length handwritten numeral string recognition
  using cascade of pca-svmnet classifiers. IEEE Access  \textbf{7},
  52024--52034 (2019). \doi{10.1109/ACCESS.2019.2911851}

\bibitem{cecotti2016active}
Cecotti, H.: Active graph based semi-supervised learning using image matching:
  application to handwritten digit recognition. Pattern Recognition Letters
  \textbf{73},  76--82 (2016)

\bibitem{Cheddad16}
Cheddad, A.: Towards query by text example for pattern spotting in historical
  documents. In: 2016 7th International Conference on Computer Science and
  Information Technology (CSIT). pp.~1--6 (2016).
  \doi{10.1109/CSIT.2016.7549479}

\bibitem{cheng2019handwritten}
Cheng, S., Shang, G., Zhang, L.: Handwritten digit recognition based on
  improved vgg16 network. In: Tenth International Conference on Graphics and
  Image Processing (ICGIP 2018). vol. 11069, p. 110693B. International Society
  for Optics and Photonics (2019)

\bibitem{cho2014learning}
Cho, K., van Merrienboer, B., Gulcehre, C., Bougares, F., Schwenk, H., Bengio,
  Y.: Learning phrase representations using rnn encoder-decoder for statistical
  machine translation. In: Conference on Empirical Methods in Natural Language
  Processing (EMNLP 2014) (2014)

\bibitem{choi2001segmentation}
Choi, S.M., Oh, I.S.: A segmentation-free recognition of handwritten touching
  numeral pairs using modular neural network. International journal of pattern
  recognition and artificial intelligence  \textbf{15}(06),  949--966 (2001)

\bibitem{CILIA2020137}
Cilia, N., {De Stefano}, C., Fontanella, F., Marrocco, C., Molinara, M.,
  {Scotto Di Freca}, A.: An end-to-end deep learning system for medieval writer
  identification. Pattern Recognition Letters  \textbf{129},  137--143 (2020).
  \doi{https://doi.org/10.1016/j.patrec.2019.11.025},
  \url{https://www.sciencedirect.com/science/article/pii/S0167865519303460}

\bibitem{cimpoi2015deep}
{Cimpoi}, M., {Maji}, S., {Vedaldi}, A.: Deep filter banks for texture
  recognition and segmentation. In: 2015 IEEE Conference on Computer Vision and
  Pattern Recognition (CVPR). pp. 3828--3836 (2015).
  \doi{10.1109/CVPR.2015.7299007}

\bibitem{ciresan2008avoiding}
Ciresan, D.: Avoiding segmentation in multi-digit numeral string recognition by
  combining single and two-digit classifiers trained without negative examples.
  In: 2008 10th International Symposium on Symbolic and Numeric Algorithms for
  Scientific Computing. pp. 225--230. IEEE (2008)

\bibitem{diem2014icfhr}
Diem, M., Fiel, S., Kleber, F., Sablatnig, R., Saavedra, J.M., Contreras, D.,
  Barrios, J.M., Oliveira, L.S.: Icfhr 2014 competition on handwritten digit
  string recognition in challenging datasets (hdsrc 2014). In: 2014 14th
  International Conference on Frontiers in Handwriting Recognition. pp.
  779--784. IEEE (2014)

\bibitem{Dutta2018}
Dutta, K., Krishnan, P., Mathew, M., Jawahar, C.V.: Improving cnn-rnn hybrid
  networks for handwriting recognition. In: 2018 16th International Conference
  on Frontiers in Handwriting Recognition (ICFHR). pp. 80--85 (Aug 2018).
  \doi{10.1109/ICFHR-2018.2018.00023}

\bibitem{gao2015deep}
Gao, B.B., Wei, X.S., Wu, J., Lin, W.: Deep spatial pyramid: The devil is once
  again in the details. arXiv preprint arXiv:1504.05277  (2015)

\bibitem{Gao2020}
{Gao}, Y., {Mishchenko}, Y., {Shah}, A., {Matsoukas}, S., {Vitaladevuni}, S.:
  Towards data-efficient modeling for wake word spotting. In: ICASSP 2020 -
  2020 IEEE International Conference on Acoustics, Speech and Signal Processing
  (ICASSP). pp. 7479--7483 (2020). \doi{10.1109/ICASSP40776.2020.9053313}

\bibitem{ghodrati2017deepproposal}
Ghodrati, A., Diba, A., Pedersoli, M., Tuytelaars, T., Gool, L.V.:
  Deepproposals: Hunting objects and actions by cascading deep convolutional
  layers. International Journal of Computer Vision  \textbf{124}(2),  115–131
  (2017). \doi{10.1007/s11263-017-1006-x}

\bibitem{gilloux2014}
Gilloux, M.: Document analysis in postal applications and check processing.
  Handbook of Document Image Processing and Recognition p. 705–747 (2014).
  \doi{10.1007/978-0-85729-859-1\_22}

\bibitem{Hochuli2020}
{Hochuli}, A.G., {Britto}, A.S., {Barddal}, J.P., {Sabourin}, R., {Oliveira},
  L.E.S.: An end-to-end approach for recognition of modern and historical
  handwritten numeral strings. In: 2020 International Joint Conference on
  Neural Networks (IJCNN). pp.~1--8 (2020).
  \doi{10.1109/IJCNN48605.2020.9207468}

\bibitem{Hochuli2020b}
Hochuli, A.G., {Britto Jr}, A.S., Saji, D.A., Saavedra, J.M., Sabourin, R.,
  Oliveira, L.S.: A comprehensive comparison of end-to-end approaches for
  handwritten digit string recognition. Expert Systems with Applications
  \textbf{165},  114196 (2021).
  \doi{https://doi.org/10.1016/j.eswa.2020.114196},
  \url{http://www.sciencedirect.com/science/article/pii/S0957417420309271}

\bibitem{hochuli2018handwritten}
Hochuli, A.G., Oliveira, L.S., Britto~Jr, A., Sabourin, R.: Handwritten digit
  segmentation: Is it still necessary? Pattern Recognition  \textbf{78},  1--11
  (2018)

\bibitem{kusetogullari2019ardis}
Kusetogullari, H., Yavariabdi, A., Cheddad, A., Grahn, H., Hall, J.: {ARDIS: A}
  swedish historical handwritten digit dataset. Neural Computing and
  Applications pp. 1--14 (2019)

\bibitem{lecun1998gradient}
LeCun, Y., Bottou, L., Bengio, Y., Haffner, P.: Gradient-based learning applied
  to document recognition. Proceedings of the IEEE  \textbf{86}(11),
  2278--2324 (1998)

\bibitem{matan1992multi}
Matan, O., Burges, C.J., LeCun, Y., Denker, J.S.: Multi-digit recognition using
  a space displacement neural network. In: Advances in neural information
  processing systems. pp. 488--495 (1992)

\bibitem{Flor2020}
Neto, A.F.D.S., Bezerra, B.L.D., Lima, E.B., Toselli, A.H.: Hdsr-flor: A robust
  end-to-end system to solve the handwritten digit string recognition problem
  in real complex scenarios. IEEE Access  \textbf{8},  208543--208553 (2020)

\bibitem{Neudecker2019}
Neudecker, C., Baierer, K., Federbusch, M., Boenig, M., W\"{u}rzner, K.M.,
  Hartmann, V., Herrmann, E.: Ocr-d: An end-to-end open source ocr framework
  for historical printed documents. In: Proceedings of the 3rd International
  Conference on Digital Access to Textual Cultural Heritage. p. 53–58.
  DATeCH2019, Association for Computing Machinery, New York, NY, USA (2019).
  \doi{10.1145/3322905.3322917}, \url{https://doi.org/10.1145/3322905.3322917}

\bibitem{Palm2019}
{Palm}, R.B., {Laws}, F., {Winther}, O.: Attend, copy, parse end-to-end
  information extraction from documents. In: 2019 International Conference on
  Document Analysis and Recognition (ICDAR). pp. 329--336 (2019).
  \doi{10.1109/ICDAR.2019.00060}

\bibitem{ribas2013handwritten}
Ribas, F.C., Oliveira, L., Britto, A., Sabourin, R.: Handwritten digit
  segmentation: a comparative study. International Journal on Document Analysis
  and Recognition (IJDAR)  \textbf{16}(2),  127--137 (2013)

\bibitem{RoyVajda2005}
{Roy}, K., {Vajda}, S., {Pal}, U., {Chaudhuri}, B.B., {Belaid}, A.: A system
  for indian postal automation. In: Eighth International Conference on Document
  Analysis and Recognition (ICDAR'05). pp. 1060--1064 Vol. 2 (2005).
  \doi{10.1109/ICDAR.2005.259}

\bibitem{Schuster1997}
{Schuster}, M., {Paliwal}, K.K.: Bidirectional recurrent neural networks. IEEE
  Transactions on Signal Processing  \textbf{45}(11),  2673--2681 (Nov 1997).
  \doi{10.1109/78.650093}

\bibitem{shi2017}
Shi, B., Bai, X., Yao, C.: An end-to-end trainable neural network for
  image-based sequence recognition and its application to scene text
  recognition. IEEE Transactions on Pattern Analysis and Machine Intelligence
  \textbf{39}(11),  2298--2304 (nov 2017). \doi{10.1109/TPAMI.2016.2646371}

\bibitem{shi2016end}
Shi, B., Bai, X., Yao, C.: An end-to-end trainable neural network for
  image-based sequence recognition and its application to scene text
  recognition. IEEE transactions on pattern analysis and machine intelligence
  \textbf{39}(11),  2298--2304 (2016)

\bibitem{simonyan2015very}
Simonyan, K., Zisserman, A.: Very deep convolutional networks for large-scale
  image recognition. In: International Conference on Learning Representations
  (2015)

\bibitem{vellasques2008filtering}
Vellasques, E., Oliveira, L.S., Britto~Jr, A., Koerich, A.L., Sabourin, R.:
  Filtering segmentation cuts for digit string recognition. Pattern Recognition
   \textbf{41}(10),  3044--3053 (2008)

\bibitem{Voigtlaender2016}
Voigtlaender, P., Doetsch, P., Ney, H.: Handwriting recognition with large
  multidimensional long short-term memory recurrent neural networks. In: 2016
  15th International Conference on Frontiers in Handwriting Recognition
  (ICFHR). pp. 228--233 (Oct 2016). \doi{10.1109/ICFHR.2016.0052}

\bibitem{wei2017selective}
Wei, X.S., Luo, J.H., Wu, J., Zhou, Z.H.: Selective convolutional descriptor
  aggregation for fine-grained image retrieval. Trans. Img. Proc.
  \textbf{26}(6),  2868–2881 (Jun 2017). \doi{10.1109/TIP.2017.2688133},
  \url{https://doi.org/10.1109/TIP.2017.2688133}

\bibitem{wen2016learning}
Wen, W., Wu, C., Wang, Y., Chen, Y., Li, H.: Learning structured sparsity in
  deep neural networks. In: Proceedings of the 30th International Conference on
  Neural Information Processing Systems. p. 2082–2090. NIPS'16, Curran
  Associates Inc., Red Hook, NY, USA (2016)

\end{thebibliography}
\bibliographystyle{splncs04}

\end{document}